# Anomaly Detection of Particle Orbit in Accelerator using LSTM Deep Learning Technology

Zhiyuan Chen, Wei Lu, Radhika Bhong, Yimin Hu, Brian Freeman, Adam Carpenter

*Abstract*— A stable, reliable, and controllable orbit lock system is crucial to an electron (or ion) accelerator because the beam orbit and beam energy instability strongly affect the quality of the beam delivered to experimental halls. Currently, when the orbit lock system fails operators must manually intervene. This paper develops a Machine Learning based fault detection methodology to identify orbit lock anomalies and notify accelerator operations staff of the off-normal behavior. Our method is unsupervised, so it does not require labeled data. It uses Long-Short Memory Networks (LSTM) Auto Encoder to capture normal patterns and predict future values of monitoring sensors in the orbit lock system. Anomalies are detected when the prediction error exceeds a threshold. We conducted experiments using monitoring data from Jefferson Lab's Continuous Electron Beam Accelerator Facility (CEBAF). The results are promising: the percentage of real anomalies identified by our solution is 68.6%-89.3% using monitoring data of a single component in the orbit lock control system. The accuracy can be as high as 82%.

*Index Terms*—neural networks, fault diagonosis

## I. INTRODUCTION

Electron orbit control has been demonstrated to be a fundamental part in the operation of high-energy/nuclear physics accelerators, such as Jefferson Lab's Continuous Electron Beam Accelerator Facility (CEBAF), and Electron-ion Collider (EIC) to be built in the future. The technique requirements on the electron beam orbit location at Jefferson Lab are extremely challenging: the system provides ±100 µm accuracy within a position range of ±5 mm for current between 400 nA and 90µA.

The 12 GeV CEBAF accelerator at Jefferson Lab is arranged in a five-pass racetrack configuration, with two Linacs joined by ten 180° magnetic transport arcs with a radius of 80 m, as shown in **Figure 1**. The source of the injector is a strained GaAs cathode providing a polarized beam of 90% polarization and maximum current of 105 µA. The electron beam from the source is first accelerated to 123 MeV, and then is injected into the north linac. Each linac consists of 25 RF cryomodules, and each module contains 8 superconducting niobium cavities. The liquid helium produced at the Central Helium Liquefier (CHL) keeps the accelerator at 2 K. The nominal gain of each linac is designed to be 1.1 GeV. Quadruple and dipole magnets in each arc provide the field which focuses and steers the beam as it passes through each arc. CEBAF is a multi-pass machine that has an independent transport arc for each pass.

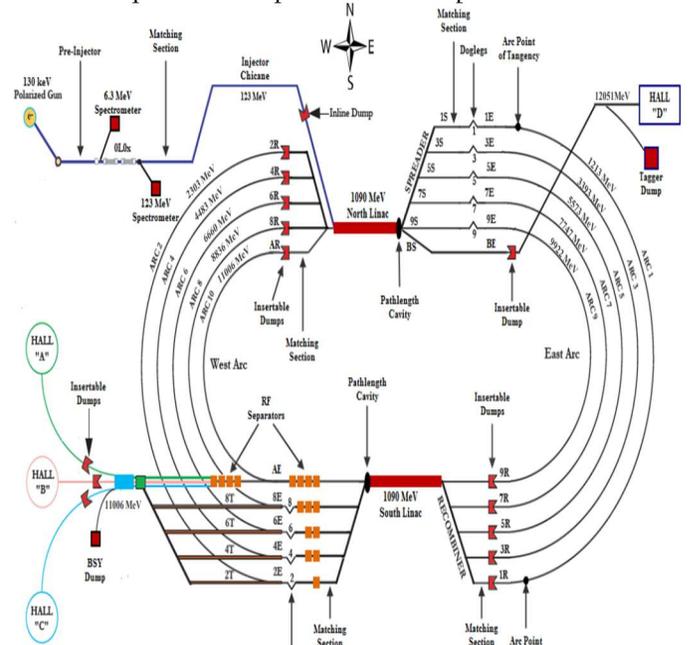

Figure 1: CEBAF Overview

A stable, reliable, and controllable orbit lock system is crucial to an electron (or ion) accelerator because the beam orbit and beam energy instability strongly affect the quality of the beam delivered to experimental Halls. Currently, when the orbit lock system fails, operators must manually intervene. For example, sufficient degradation of a magnet power supply downstream of an orbit lock will cause an orbit anomaly. Hundreds of such magnets exist in CEBAF, and ten such magnet incidents in 4Q FY20 which caused a total of 7.3 hours of beam downtime.

Neural networks have been used for control and modeling of particle accelerators [1-4], image diagnosis [5], and optimizing parameters [6, 7].

Anomaly detection has been applied to the accelerator operation where ML can significantly improve the safety and efficiency [8, 9]. However, both works require feature engineering. This paper uses deep-learning techniques that do not require feature engineering. Recent efforts have been aimed

This work was supported in part by the U.S. Department of Energy under Grant DE-SC0022438.

Zhiyuan Chen is now with the Department of Information Systems, University of Maryland Baltimore County, Baltimore, MD 21250 USA (e-mail: zhchen@umbc.edu).

Wei Lu and Yimin Hu are with Raytum Photonics, Sterling, VA 20166 (email: Wei.Lu@raytum-photonics.com, yimin.hu@raytum-photonics.com).

Radhika Bhong was with with the Department of Information Systems, University of Maryland Baltimore County, Baltimore, MD 21250 USA (e-mail: rbhong1@umbc.edu).

Brian Freeman is with Jefferson Lab, Newport News, VA 23606 USA (e-mail: bfreeman@jlab.org).

Adam Carpenter is with Jefferson Lab, Newport News, VA 23606 USA (e-mail: adamc@jlab.org)

at understanding and predicting faulty behavior in superconducting RF cavities and magnets [10]. This is of interest due to the potentially catastrophic nature of a failure in these devices. Besides, ML has been applied to locate and isolate malfunctioning beam position monitors in the LHC, prior to application of standard optics correction procedures [11, 12]. Autoencoders are also used in [13] in particle accelerator. But the goal of that work is just to detect magnet faults. In this paper we consider all kinds of faults that cause sudden drop of beam current.

Motivated by the above considerations, the objective of this paper is to develop a ML based fault detection methodology to identify orbit lock anomalies and notify accelerator operations staff of the off-normal behavior. Improved understanding of anomalous orbit lock behavior will provide two valuable enhancements to CEBAF's operational performance. First, tighter temporal tuning of the locks will reduce the position variation on experimental targets when the orbit changes. This is an ever-present issue as the accelerator tunnel continually expands and contracts due to both daily and seasonal temperature changes. Second, prompt identification of off-normal performance by orbit lock sensors and controllers can provide an early warning that the underlying system requires support. This can improve accelerator reliability when hardware and support labor can be readied as soon as deteriorating behavior is detected, rather than waiting for outright failure.

In this paper, we are going to present our design and development of a machine-learning based anomaly detection method for orbit-lock control, using monitoring data from Jefferson Lab's Continuous Electron Beam Accelerator Facility (CEBAF). This method uses the state-of-the-art LSTM auto-encoder structure to capture normal patterns of monitoring data and detect anomalies when reconstruction error exceeds a threshold. This method is unsupervised so it does not require labeled training data. Our solution is able to identify the majority of real anomalies just using monitoring data from one component of the orbit-lock control system. The percentage of real anomalies identified by our solution is 68.6%-89.3% using monitoring data of a single component in orbit lock control system. The accuracy can be as high as 82%.

Section II describes the machine-learning based anomaly detection for orbit-lock control. Section III presents experimental results. Section IV concludes the paper.

## II. LSTM BASED ANOMALY DETECTION

In our study, we employed an LSTM (Long Short-Term Memory) autoencoder to detect typical patterns in beam monitoring data. LSTM, a recurrent neural network type, is well-known for its ability to learn long-term dependencies and its aptitude for processing time series data. A LSTM autoencoder is constructed on top of a number of LSTM units and can receive a time series as input, predict the subsequent value based on observed values in a preceding time window, and reconstruct the time series. One of the main advantages of using an LSTM autoencoder is its self-supervised nature, as we do not need to label the data because the autoencoder tries to minimize the reconstruction error.

Our solution consists of three steps: 1) extract normal data from observed orbit lock data by removing known anomalies; 2) train a LSTM autoencoder based on the extracted data to capture patterns of normal data; 3) run the autoencoder on testing data and flag an anomaly when the reconstruction error exceeds a pre-defined threshold. The rationale is that if the new data value deviates significantly from the normal pattern predicted by the LSTM autoencoder, an anomaly has likely occurred.

We first give a brief introduction of LSTM. We then describe the data set, pre-processing, and the anomaly detection step.

### A. LSTM Autoencoder Structure

An autoencoder takes input data in a time series format, where each row represents a time series from a particular monitor. Each column denotes a timestep within a sliding window of size $k$, with one sample every time unit. So if there are $m$ monitors the input is a $k$ by $m$ matrix at every point of time. The output of the autoencoder is reconstructed data as a time series as well as predicted values at the next time step. So the output is a $k$ by $m+1$ matrix.

An autoencoder consists of an encoder which compresses input to an embedding with lower dimensions followed by a decoder that reconstructs the input from the embedding. Both encoder and decoder consist of several LSTM layers where each layer has a number of LSTM units. Next we describe a LSTM unit and structure of the autoencoder.

Each LSTM unit consists of a cell, an input gate, an output gate, and a forget gate. The cell stores long-term dependency information, while the gates regulate the cell's input/output.

The first layer of the encoder consists of $k$ LSTM units. These units are connected sequentially as shown in Figure 2, and each unit receives input at time step $t$ (where $t$ ranges from 1 to $k$), and outputs a vector $h_t$.

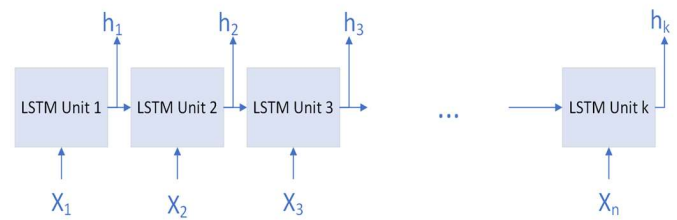

**Figure 2**. A LSTM layer with $k$ LSTM units

We utilized the Keras package [14] in Python to implement a LSTM autoencoder with the structure shown in Figure 3. For our experiments, we utilized a single layer of LSTM in the encoder. The LSTM layer takes in a time sequence with a window size of 30 seconds and three features: wiresum, x-position, and y-position of the beam monitors. This particular LSTM layer comprises of 30 LSTM units, with one dedicated to each time step. Its output generates an embedding of 64 dimensions. We used mean absolute error as loss function and the 'Adam' optimizer to optimize the input weights by comparing the prediction with the loss function.

**Figure 3** illustrates the design of a LSTM auto-encoder.

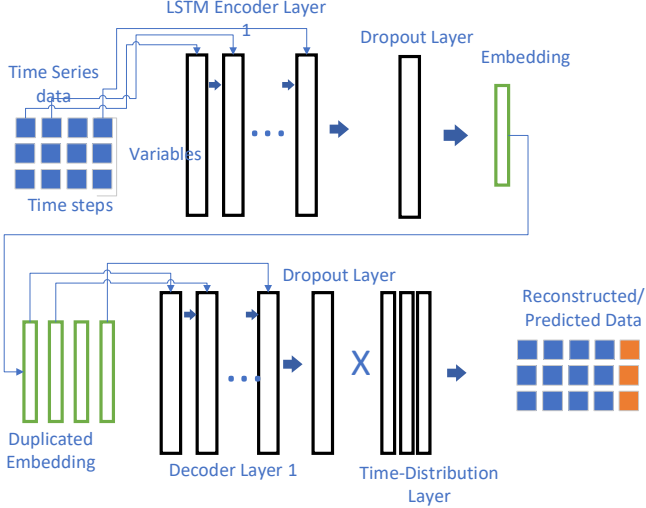

**Figure 3**. LSTM auto-encoder

The autoencoder consists of the following layers:

● Layer 1 with 64 LSTM units. It reads the input data and outputs 64 features with 30 timesteps.

● Layer 2, a dropout layer with drop out rate=0.2. This layer helps prevent overfitting. The output of this layer is the embedding vector of the input data.

● Layer 3, RepeatVector(30), replicates the feature vector 30 times. The RepeatVector layer acts as a bridge between the encoder and decoder modules. It prepares the 2D array input for the first LSTM layer in Decoder.

● Layer 4, the Decoder layer to unfold the encoding. It also consists of 64 LSTM units.

● Layer 5, dropout layer with drop out rate=0.2. Layer 4 and 5 are the mirror images of Layer 1 and Layer 2 respectively.

● Layer 6, TimeDistributed layer followed by a dense layer to generate the output of a 30 by 3 matrix, where "3" is the number of features in the input data.

### B. Data and Data Pre-Processing

We used data collected by Beam Position Monitors and Beam Loss Monitors in the injector, first Linac, and first arc at the Jefferson Lab's CEBAF accelerator (see Figure 1). These data are most closely related to the health of beams. The data is in the form of text files containing values from various monitoring sensors located near different components of CEBAF. We used data collected over three two-day period in December 2021, January 2022 and February 2022 to ensure there was enough variety in the data set. These files are first converted into time series format. 50% of the data is used for training the model, while the remaining 50% is used for testing. Because the variables are sampled at different rates, data from multiple variables is integrated by merging data from different variables into the same time unit, resampled at the second level. Since there are many missing values, a forward fill interpolation method is used to replace the missing values.

Table 1 shows properties of training and testing data sets.

We used three position monitors (IPM1A04, IPM0R01, IMP1L02). Each monitor generates three time series data: wire-sum, X- and Y-positions.

Table 1. Properties of Data Sets

| Monitor Name | Training Data Range | Training Data Size | Testing Data Range | Testing Data Size |
|---|---|---|---|---|
| IPM1A04 (first arc) | Dec 16 to Dec 18 2021 | 135782 | Dec 19 to 20 2021 | 172800 |
| IPM0R01 (injector) | Jan 21 to 23, 2022 | 161531 | Jan 24 to 25 2022 | 113916 |
| IPM1L02 (1st Linac) | Feb 4 to Feb 6, 2022 | 136067 | Feb 7 to 8 2022 | 112540 |

To decide the ground truth, we used two fault files that record beam loss faults and RF-related faults, which account for more than 90% of all types of faults. These files contain the starting timestamp of the recorded faults. However, not all faults have been recorded, so we also use a heuristic method to identify additional faults. We used a beam current file to identify periods when the beam current dropped below a given threshold. A sudden drop in beam current typically indicates abnormal behavior, including faults. So our ground truth includes both faults in the two fault files and the drops of beam current identified using the heuristic method.

To build an ML model that can capture patterns for normal data, it is important to extract normal data. The training data contains many faults that are clearly abnormal, so we removed the data values ten seconds before and after faults from the training dataset to obtain normal data values for model training. The test dataset is unchanged.

Figure 4 illustrates a sample of the beam current data, wire sum data, and X- and Y-positions. The known faults (those from the two fault files) are indicated by the vertical yellow lines and there are also instances of beam loss shown in the beam current data. It is evident from the plot that sudden changes occur in all three signals around faults and shutdowns. It can be also seen that some likely faults (indicated by sudden drops of beam current) are not included in the two files.

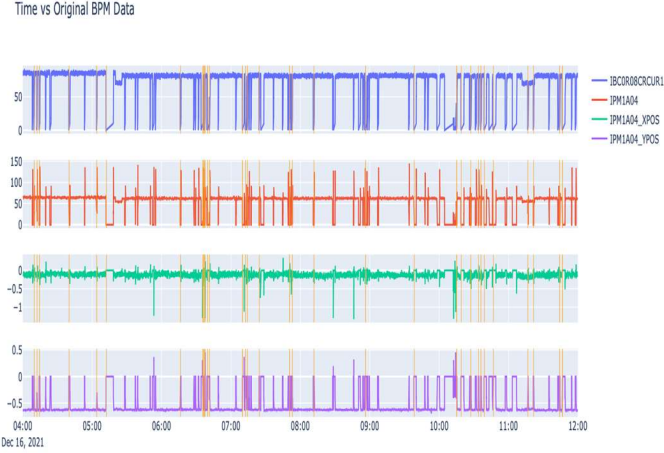

Figure 4. Sample training data and beam current data before extraction of normal data

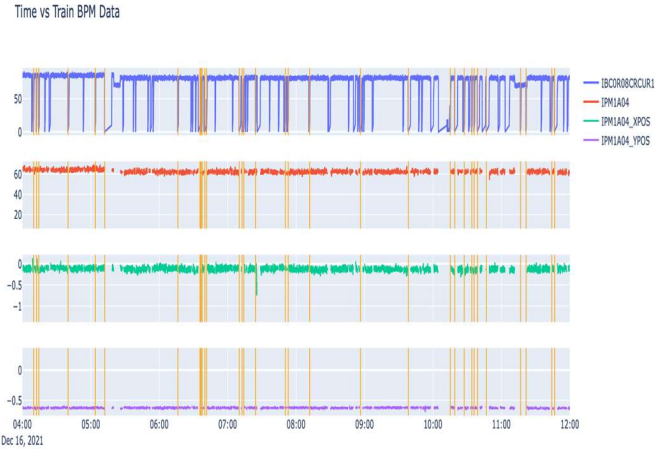

Figure 5. Sample extracted training data and beam current data

On the other hand, Figure 5 depicts the same data after extracting normal data. In this plot, there are gaps observed before or after faults or shutdowns. It is apparent that all three signals, namely wiresum, x-position, and y-position, are comparatively more stable in the normal data.

## C. Anomaly Detection Using the LSTM Auto-Encoder Model

An autoencoder will be trained using the extracted normal training data to both reconstruct the initial data and predict data at the next time step. The trained autoencoder is then applied to testing data to detect anomalies. Since the autoencoder learns to recognize normal data patterns, so the reconstruction error over normal data is expected to be low. Therefore, high reconstruction error indicates abnormal data. So the anomaly detection method employs a threshold for reconstruction error. If the reconstruction error for the current monitoring data exceeds the threshold, it signals the presence of an anomaly.

To set the threshold, we used the three-sigma rule. We calculated the mean and standard deviation of the reconstruction error on the training data and set the threshold at mean + 3 * standard deviation. We experimented with different threshold values and found that 3 sigma gave the best results on the testing data, which corresponded to approximately 15-20% of the maximum absolute error loss of the training dataset.

## III. Experimental Results

Figure 6 shows anomalies for test data for IPM1A04 in data for 19-Dec-2021 from 1:30 to 3:30AM. Anomalies are shown as dots and known faults are shown as yellow vertical lines along X-axis.

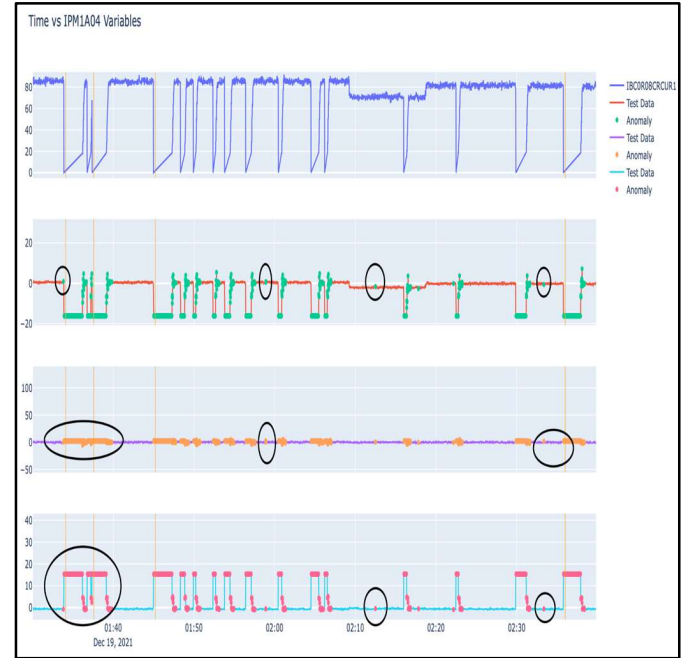

**Figure 6.** Anomalies discovered in testing data from 1:30 to 3:30AM 19-Dec-2021. The top chart is beam current (not used in training). The second chart is IPM1A04. The third is IPM1A04.XPOS, and the bottom is IPM1A04.YPOS. Yellow lines are known faults. Dots are anomalies.

We can see that the anomalies are detected either right before, during, or right after a fault or shutdown (see the first circle in the chart for three monitoring sensors). Our method is able to detect almost all known faults (those in yellow lines). In addition, our method also detects small but unusual fluctuations of beam current. For example, the second, third, the fourth circles in monitoring data correspond to unusual but less dramatic fluctuations of beam current. Such fluctuation may indicate a future fault. There are sudden drops of beam current about a minute or a couple of minutes after the unusual fluctuations indicated by these anomalies.

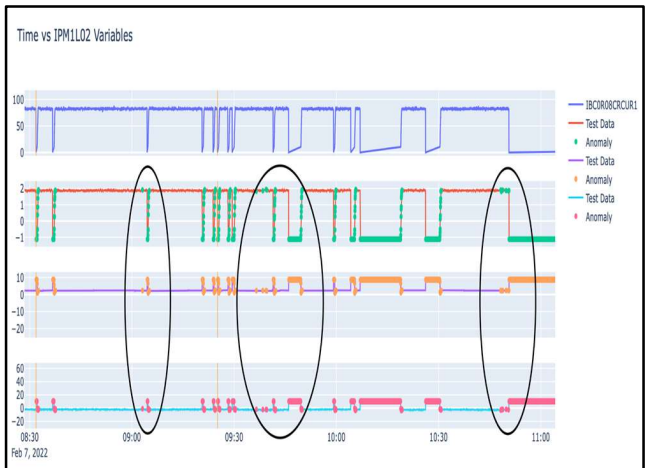

**Figure 7**. Anomalies discovered in testing data from 8:30 to 11 am Feb 7, 2022. The top chart is beam current. The second chart is IPM1L02. The third is IPM1L02.XPOS, and the bottom is IPM1L02.YPOS. Yellow lines are known faults. Dots are anomalies.

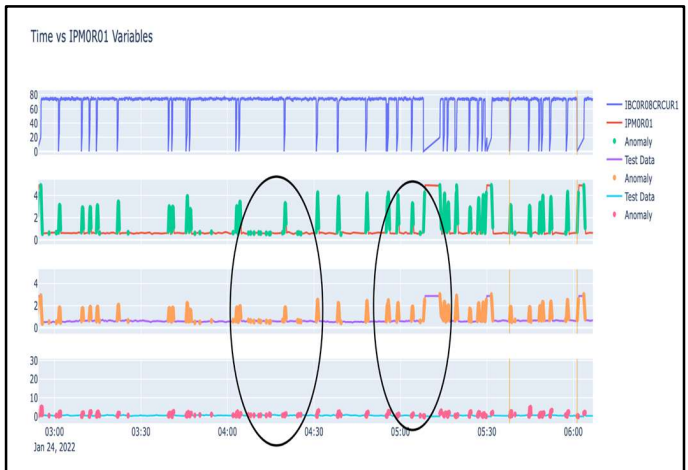

Figure 8: Anomalies discovered in testing data from 3 am to 6 am Jan 24, 2022. The top chart is beam current. The second chart is IPM0R01. The third is IPM0R01.XPOS, and the bottom is IPM0R01.YPOS. Yellow lines are known faults. Dots are anomalies.

Figure 7 shows results for 8:30 am to 11 am Feb 7, 2022 and Figure 8 shows results for 3 am to 6 am January 24, 2022. We have similar observations: our solutions detect anomalies right before, during, and right after faults or shutdowns. For example, in the first circle in Figure 7, one anomaly was detected within a minute before a sudden drop of beam current, which is likely a fault. A few other anomalies were detected also during the fault and when the beam current was restored. So these results showed that our method was able to detect imminent faults.

Table 2 shows precision, recall, accuracy, and F1 score for these three testing data sets. We consider an anomaly indicates a fault or shutdown if it is within 10 seconds before the start of the fault or shutdown.

Table 2. Performance measures for various test data

| Data sets | Precision | Recall | Accuracy | F1-Score |
|---|---|---|---|---|
| IPM1A04 | 0.26 | 0.893 | 0.521 | 0.403 |
| IPM1L02 | 0.102 | 0.703 | 0.766 | 0.178 |
| IPM1R01 | 0.581 | 0.686 | 0.822 | 0.629 |

Our findings reveal that our solution exhibits a high recall rate ranging from 0.686 to 0.893, indicating that our approach can successfully detect most faults or shutdowns based on monitoring data. This is a promising result, considering that we utilized monitoring data from only one component in the orbit lock control system and employed solely three variables in each dataset. If we expand our approach to more components and variables, we may be able to further enhance the recall rate.

However, our approach suffers from relatively low precision, primarily because it identifies anomalies not only before but also during or after a fault or shutdown. To address this issue, we suggest merging consecutive anomalies since they often correspond to the same fault or shutdown, similar to the approach commonly used in intrusion detection systems.

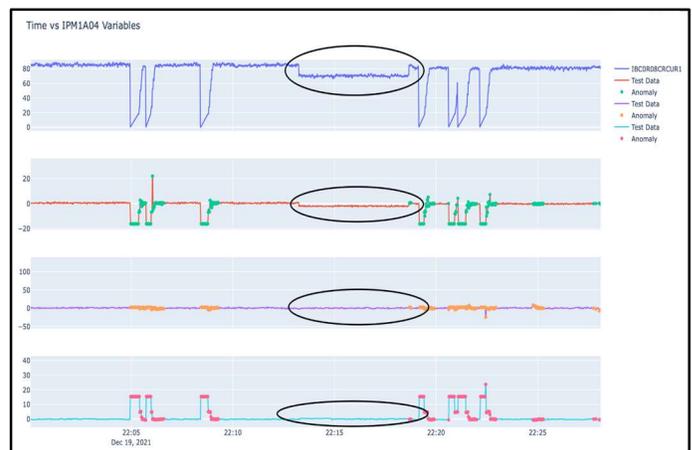

Figure 9: missed anomalies for IPM1A04 December 19, 2021

Table 2 indicates that our model still misses a few possible faults in the test data. In some cases, the model failed to detect faults or shutdowns when there were minor changes in values well before or after the actual shutdown, when the beam current dropped to zero. In such cases, the reconstruction error remained relatively low, and our method did not identify these earlier anomalies. However, in these instances, anomalies were still detected closer to the faults or shutdowns when values changed more drastically. Figure 9 shows an example of missed anomalies in detecting the shutdown for the IPM1A04 variable. There is a minor drop of beam current around 10:12 pm December 19, 2021 and then a few more drastic drops later around 10:20 pm. Our method was not able to detect the minor drop, but was able to detect more drastic drops later. So in future work, we will investigate methods that can detect less drastic changes in the data.

## IV. Conclusion

We described a machine learning approach to detect anomalies in an orbit lock system of an electron accelerator. Our solution is unsupervised so it does not require labeled data. The proposed approach uses a LSTM autoencoder to capture normal patterns of monitoring data and then detects an anomaly when the error for predicted value exceeds a threshold. Experimental results using monitors for just one component of Jefferson Lab's Continuous Electron Beam Accelerator Facility (CEBAF) show the proposed approach detects 68.6%-89.3% of true faults. This clearly shows the potential of the proposed approach.

There are a few limitations of the proposed solution: 1) our study only uses monitors in one component of the orbit lock system in each data set; 2) the solution has low precision because it also detects anomalies during the fault or in the recovery stage, which are less useful for the operators; 3) some possible indicator of faults such as less dramatic fluctuations in the monitoring data are not detected as faults due to relative small reconstruction errors. In future we will address the first issue by applying our approach to monitoring data from more components. We will address the second issue by merging consecutive anomalies that correspond to the same fault. To address the third issue, we plan to investigate methods that can detect less dramatic fluctuations, possibly using supervised learning methods.